\documentclass[11pt,a4paper]{article}
\usepackage[hyperref]{acl2019}
\usepackage{times}
\usepackage{latexsym}

\usepackage{url}

\usepackage{amsmath}
\usepackage{adjustbox}
\usepackage{CJKutf8}
\usepackage{tablefootnote}
\usepackage{graphicx}
\usepackage{subcaption}
\usepackage{url}
\usepackage{wrapfig}
\usepackage{sidecap}
\usepackage{physics}
\usepackage{mdwlist}
\usepackage{enumitem}
\setlist[enumerate]{itemsep=1mm}

\newif\ifcomment\commentfalse

\usepackage{framed}
\usepackage{mdwlist}
\usepackage{latexsym}
\usepackage{xcolor}
\usepackage{nicefrac}
\usepackage{booktabs}
\usepackage{amsfonts}
\usepackage[T1]{fontenc}
\usepackage{bold-extra}
\usepackage{amsmath}
\usepackage{dsfont}
\usepackage{amssymb}
\usepackage{bm}
\usepackage{graphicx}
\usepackage{mathtools}
\usepackage{microtype}
\usepackage{multirow}
\usepackage{multicol}
\usepackage{latexsym,comment}

\newcommand{\gem}[1]{\mbox{\textsc{gem}}}
\newcommand{\abr}[1]{\textsc{#1}}

\newcommand{\abrcamel}[2]{{\textsc #1}{\small{#2}}}

\newcommand{\email}[1]{ {\small \href{mailto://#1}{\texttt{#1} }  }}

\newcommand{\ind}[1]{\mathds{1}\left[ #1 \right] }

\newcommand{\hidetext}[1]{}
\newcommand{\ignore}[1]{}

\ifcomment
\newcommand{\pinaforecomment}[3]{\colorbox{#1}{\parbox{.8\linewidth}{#2: #3}}}
\else
\newcommand{\pinaforecomment}[3]{}
\fi

\newcommand{\jbgcomment}[1]{\pinaforecomment{red}{JBG}{#1}}

\newcommand{\smallurl}[1]{ \begin{tiny}\url{#1}\end{tiny}}

\definecolor{lightblue}{HTML}{3cc7ea}
\definecolor{CUgold}{HTML}{CFB87C}
\definecolor{grey}{rgb}{0.95,0.95,0.95}
\definecolor{ceil}{rgb}{0.57, 0.63, 0.81}

\aclfinalcopy 
\def\aclpaperid{454} 

\newcommand{\figfile}[1]{2019_acl_modularity/figures/#1}
\newcommand{\autofig}[1]{2019_acl_modularity/auto_fig/#1}

\title{A Resource-Free Evaluation Metric for Cross-Lingual Word Embeddings Based on Graph Modularity}

\author{Yoshinari Fujinuma \\
        Computer Science \\
        University of Colorado \\
        \email{fujinumay@gmail.com}
        \And
        Jordan Boyd-Graber \\
        \abr{cs}, iSchool, \abr{umiacs}, \abr{lsc} \\
        University of Maryland \\
        \email{jbg@umiacs.umd.edu}
        \And
        Michael J. Paul\\
        Information Science\\
        University of Colorado \\
        \email{mpaul@colorado.edu}
        }

\date{}

\begin{document}
\maketitle
\begin{abstract}

Cross-lingual word embeddings encode the meaning of words from different
languages into a shared low-dimensional space.
An important requirement for many downstream tasks is that word
similarity should be independent of language---i.e., word vectors
within one language should not be more similar to each other than to
words in another language.
We measure this characteristic using {\it modularity}, a
network measurement that measures the strength of clusters in a graph.
Modularity has a moderate to strong correlation with three downstream
tasks, even though modularity is based only on the structure of
embeddings and does not require any external resources.
We show through experiments that modularity can serve as an intrinsic validation metric to
improve unsupervised cross-lingual word embeddings, 
particularly on distant language pairs in low-resource settings.\footnote{Our
code is at \url{https://github.com/akkikiki/modularity_metric}}



\end{abstract}

\section{Introduction}

The success of monolingual word embeddings in natural language
processing~\citep{NIPS2013_5021} has motivated extensions to
cross-lingual settings.
Cross-lingual word embeddings---where multiple languages share a single
distributed representation---work well for
classification~\citep{klementiev-titov-bhattarai:2012:PAPERS,ammar2016massively}
and machine
translation~\citep{lample2018unsupervised,artetxe2018unsupervised},
even with few bilingual pairs~\citep{artetxe-labaka-agirre:2017:Long}
or no supervision at all~\cite{adv_bli,lample2018word,self_learn}.

Typically the quality of cross-lingual word embeddings is measured with respect to
how well they improve a downstream task.
However, sometimes it is not possible to evaluate embeddings for a specific downstream task,
for example a future task that does not yet have data
or on a rare language that does not have resources to support traditional evaluation.
In such settings, it is useful to have an {\em intrinsic} evaluation
metric: a metric that looks at the embedding space itself to know whether
the embedding is good \emph{without} resorting to an
extrinsic task.
While extrinsic tasks are the ultimate arbiter of
  whether cross-lingual word embeddings work, intrinsic metrics are useful for low-resource
  languages where one often lacks the annotated data that would make an
  extrinsic evaluation possible.

However, few intrinsic measures exist for cross-lingual word embeddings,
and those that do exist require external linguistic resources (e.g.,
sense-aligned corpora in \citet{ammar2016massively}).
The requirement of language resources makes this approach limited or impossible for low-resource languages,
which are the languages where intrinsic evaluations are most needed.
Moreover, requiring language resources can bias the evaluation toward words in the
resources rather than evaluating the embedding space as a whole.

Our solution involves a graph-based metric that considers the characteristics of the embedding space
without using linguistic resources.
To sketch the idea, imagine a cross-lingual word embedding space where
it is possible to draw a hyperplane that separates all word vectors in one language from all vectors in another.
Without knowing anything about the languages, it is easy to see that this is a problematic embedding:
the representations of the two languages are in distinct parts of the space rather than using a shared space.
While this example is exaggerated, this characteristic where vectors are clustered by language often appears
within smaller {neighborhoods} of the embedding space,
we want to discover these clusters.

To measure how well word embeddings are mixed across languages, we draw on concepts
from network science.
Specifically, some cross-lingual word embeddings are
\emph{modular} by language: {\bf vectors in one language are
  consistently closer to each other than vectors in another language}
(Figure~\ref{fig:assort_disassort}).
When embeddings are modular, they often fail on
downstream tasks (Section~\ref{sec:diagnosis}).

\begin{figure}[t]
 \centering
  \begin{subfigure}[t]{0.52\linewidth}
 \centering
     {\includegraphics[width=0.95\linewidth]{\figfile{top_cos_sim_graph_enlarged.pdf}}}
     \caption{low modularity}
  \end{subfigure}
  \begin{subfigure}[t]{0.46\linewidth}
 \centering
     {\includegraphics[width=0.95\linewidth]{\figfile{bottom_cos_sim_graph_enlarged.pdf}}}
     \caption{\label{fig:assort}high modularity}
  \end{subfigure}
     \caption{\label{fig:assort_disassort} An example of a low
       modularity (languages mixed) and high
       modularity cross-lingual word embedding lexical
       graph using $k$-nearest neighbors of ``eat'' (left) and
       ``firefox'' (right) in English and Japanese.
     }
\end{figure}

Modularity is a concept from network theory
(Section~\ref{sec:modularity}); because network theory is applied to
graphs, we turn our word embeddings into a graph by connecting
nearest-neighbors---based on vector similarity---to each other.
Our hypothesis is that {\em modularity will predict how useful the
  embedding is} in downstream tasks; low-modularity embeddings should
work better.

We explore the relationship between modularity and three downstream
tasks (Section~\ref{sec:experiment}) that use cross-lingual word embeddings differently: (i)
cross-lingual document classification; (ii) bilingual lexical
induction in Italian, Japanese, Spanish, and Danish; and (iii)
low-resource document retrieval in Hungarian and Amharic, finding
moderate to strong negative correlations between modularity and performance.
Furthermore, using modularity as a validation metric
(Section~\ref{sec:validation}) makes \abr{muse}~\cite{lample2018word}, an unsupervised model,
more robust on distant language pairs.
Compared to other existing intrinsic evaluation metrics, modularity
captures complementary properties and is more predictive of downstream
performance despite needing no external resources (Section~\ref{sec:empirical}).

\section{Background: Cross-Lingual Word Embeddings and their Evaluation}
\label{sec:diagnosis}

There are many approaches to training cross-lingual word embeddings.
This section reviews the embeddings we consider in this paper,
along with existing work on evaluating those embeddings.

\subsection{Cross-Lingual Word Embeddings}

We focus on methods that learn a cross-lingual vector space through a post-hoc mapping between independently constructed
monolingual embeddings~\citep{DBLP:journals/corr/MikolovLS13,vulic-korhonen-acl:2016c}.
Given two separate monolingual embeddings and a bilingual seed
lexicon, a projection matrix can map translation pairs in a given
bilingual lexicon to be near each other in a shared embedding space.
A key assumption is that cross-lingually coherent words have ``similar
geometric arrangements''~\citep{DBLP:journals/corr/MikolovLS13}
in the embedding space, enabling ``knowledge transfer between
languages''~\citep{DBLP:journals/corr/Ruder17}. 

We focus on mapping-based approaches for two reasons.  First, these approaches are
applicable to low-resource languages because they do not requiring
large bilingual dictionaries or parallel corpora
\citep{artetxe-labaka-agirre:2017:Long,lample2018word}.\footnote{\citet{DBLP:journals/corr/Ruder17}
  offers detailed discussion on alternative approaches. }
Second, this focus separates the word embedding task from the cross-lingual
mapping, which allows us to focus on evaluating the specific
multilingual component in Section~\ref{sec:experiment}.

\subsection{Evaluating Cross-Lingual Embeddings}

Most work on evaluating cross-lingual embeddings
 focuses on extrinsic evaluation of downstream tasks~\citep{upadhyay-EtAl:2016,glavas2019properly}.
However, intrinsic evaluations are crucial since many low-resource languages
lack annotations needed for downstream tasks.
Thus, our goal is to develop an intrinsic measure that correlates
with downstream tasks without using any external resources.
This section summarizes existing work on intrinsic methods of evaluation for cross-lingual embeddings.

One widely used intrinsic measure for evaluating the coherence of
monolingual embeddings is \textsc{qvec}~\citep{qvec:enmlp:15}. 
\citet{ammar2016massively} extend \textsc{qvec} by using 
canonical correlation analysis (\textsc{qvec-cca})
to make the scores comparable across
embeddings with different dimensions.
However, while both \textsc{qvec} and \textsc{qvec-cca} can be
extended to cross-lingual word embeddings, they are limited: they require external annotated corpora.  
This is problematic in cross-lingual settings since this requires
annotation to be consistent across
languages~\citep{ammar2016massively}.

Other internal metrics do not require external resources, but those consider only part of the embeddings.
\citet{lample2018word} and \citet{self_learn} use a validation metric
that calculates similarities of cross-lingual neighbors to conduct
model selection.
Our approach differs in that we consider whether cross-lingual nearest neighbors 
are {\it relatively closer} than intra-lingual nearest neighbors.

\citet{eigenval_sim} use the similarities of intra-lingual
neighbors and compute graph similarity between two monolingual
lexical subgraphs built by subsampled words in a bilingual lexicon.
They further show that the resulting graph similarity has a high
correlation with bilingual lexical induction on \abr{muse}~\citep{lample2018word}.
However, their graph similarity still only uses intra-lingual similarities
but not cross-lingual similarities. 

These existing metrics are limited by either requiring external resources
or considering only part of the embedding structure (e.g., intra-lingual but not cross-lingual neighbors).
In contrast, our work develops an intrinsic metric which
 is highly correlated with multiple downstream tasks but does not require external resources,
and considers both intra- and cross-lingual neighbors. 

\paragraph{Related Work}
A related line of work is the intrinsic evaluation measures of probabilistic
topic models, which are another low-dimensional representation of
words similar to word embeddings.  Metrics based on word co-occurrences
have been developed for measuring the monolingual coherence of topics
\citep{NewmanLGB10,MimnoWTLM11,LauNB14}. Less work has studied
evaluation of cross-lingual topics \citep{MimnoWNSM09}. Some
researchers have measured the overlap of direct translations across
topics~\citep{Boyd-Graber:Blei-2009}, while \citet{Hao18mltm} propose
a metric based on co-occurrences across languages that is more general
than direct translations.

\section{Approach: Graph-Based Diagnostics for Detecting Clustering by Language}
\label{sec:modularity}

This section describes our graph-based approach to measure the intrinsic quality of a cross-lingual embedding space.

\subsection{Embeddings as Lexical Graphs}

We posit that we can understand the quality of cross-lingual
embeddings by analyzing characteristics of a lexical
graph~~\citep{pelevina-etal-2016-making,hamilton-EtAl:2016:EMNLP2016}.  The lexical graph has
words as nodes and edges weighted by their similarity in the embedding space.
Given a pair of words $(i, j)$ and associated word vectors~$(v_i,
v_j)$, we compute the similarity between two
words by their vector similarity.  We encode this similarity in a weighted adjacency matrix~$A$: $A_{ij} \equiv \max(0, \text{cos\_sim}(v_i,
v_j))$.
However, nodes are only connected to their $k$-nearest neighbors
(Section~\ref{sec:k_sensitivity} examines the sensitivity to $k$); all
other edges become zero.
Finally, each node $i$ has a label $g_i$ indicating the word's language.

\subsection{Clustering by Language}

\begin{figure}[!tb]
 \centering
     {\includegraphics[width=0.95\linewidth]{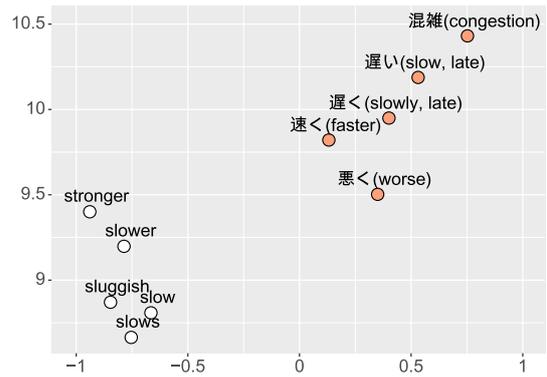}}
  \vspace{-1ex}
     \caption{ Local t-\abr{sne}~\citep{t-sne} of an
       \textsc{en}-\textsc{ja} cross-lingual word embedding, 
       which shows an example of ``clustering by language''. 
        }
\label{fig:tsne}
\end{figure}

We focus on a phenomenon that we call ``clustering by language'',
when word vectors in the embedding space tend to be more similar to words in the same language than words in the other.
For example in Figure~\ref{fig:tsne}, 
 the intra-lingual nearest neighbors of ``slow''
have higher
similarity in the embedding space than semantically related cross-lingual words.
This indicates that words are represented differently across the two languages,
thus our hypothesis is that clustering by language degrades the quality of cross-lingual embeddings when used in downstream tasks. 

\subsection{Modularity of Lexical Graphs}

With a labeled graph, we can now ask whether the graph is
\emph{modular}~\cite{newman2010networks}.
 In a cross-lingual
lexical graph, modularity is the degree to which words are more
similar to words in the \emph{same} language than to words in a
\emph{different} language.  This is undesirable, because the
representation of words is not transferred across languages. 
If the nearest neighbors of the words are instead within the same language,
then the languages are not mapped into the cross-lingual space
consistently.  In our setting, the language $l$ of each word defines
its group, and \emph{high} modularity indicates embeddings are more
similar \emph{within} languages than \emph{across}
languages~\citep{newman2003mixing,newman2004finding}. 
In other words, good embeddings should have \emph{low} modularity.

Conceptually, the modularity of a lexical graph is the difference
between the proportion of edges in the graph that connect two nodes
from the same language and the {\em expected} proportion of such
edges in a randomly connected lexical graph.
If edges were random, the number of edges starting from node~$i$ within the same language would be the degree of node~$i$,~$d_i
= \sum_j A_{ij}$ for a weighted graph, following \citet{PhysRevE.70.056131}, times the proportion of words in that language.  Summing over all nodes gives the expected number of edges within a language,
\begin{equation}
a_{l} = \frac{1}{2m} \sum_{i} d_{i} \ind{g_i=l},
\end{equation}
where $m$ is the number of edges, $g_i$ is the label of node $i$, and $\ind{\cdot}$ is an indicator function
that evaluates to $1$ if the argument is true and $0$ otherwise. 

Next, we count the fraction of edges $e_{ll}$ that 
connect words of the same language:\begin{equation}
e_{ll} = \frac{1}{2m} \sum_{ij} A_{ij} \ind{g_i=l} \ind{g_j=l}.
\end{equation}
Given $L$ different languages, we calculate overall modularity~$Q$ by
taking the difference between $e_{ll}$ and $a_l^2$ for all languages:\begin{equation}
\label{unnorm_modularity}
Q = \sum_{l = 1}^L (e_{ll} - a_l^2).
\end{equation}
Since $Q$ does not necessarily have a maximum value of $1$, 
we normalize modularity:\begin{equation}
Q_{norm} = \frac{Q}{Q_{max}}, \textrm{where} \,\, Q_{max} = 1 - \sum_{l = 1}^L (a_l^2). 
\end{equation}The higher the modularity, the more words from the same language
appear as nearest neighbors.  Figure~\ref{fig:assort_disassort} shows
the example of a lexical subgraph with low modularity (left, $Q_{norm} =
0.143$) and high modularity (right, $Q_{norm}=0.672$).  In
Figure~\ref{fig:assort}, the lexical graph is modular since
``firefox'' does not encode same sense in both languages.

Our hypothesis is that cross-lingual word embeddings with lower modularity
will be more successful in downstream tasks.
If this hypothesis holds, then modularity could be a useful
metric for cross-lingual evaluation.

\section{Experiments: Correlation of Modularity with Downstream Success}
\label{sec:experiment}

We now investigate whether
modularity can predict the effectiveness of cross-lingual word embeddings
on three downstream tasks: (i) cross-lingual document classification,
(ii) bilingual lexical induction, and (iii) document retrieval in
low-resource languages.
If modularity correlates with task
performance, it can characterize embedding quality.

\subsection{Data}

To investigate the relationship between embedding effectiveness and
modularity, we explore five different cross-lingual word embeddings
on six language pairs (Table~\ref{tab:stats_mono}).

\begin{table}[!tb]
\centering
\small
 \begin{tabular}{llr}
\bf Language  & \bf Corpus                      & \bf Tokens        \\ \hline
     English (\textsc{en})   & News             & 23M             \\
     Spanish (\textsc{es})   & News             & 25M             \\
     Italian (\textsc{it})   & News             & 23M             \\
     Danish (\textsc{da})    & News             & 20M             \\
     Japanese (\textsc{ja})  & News             & 28M             \\
     Hungarian (\textsc{hu}) & News             & 20M             \\
     Amharic (\textsc{am})   & \textsc{lorelei} & 28M             \\
 \hline
 \end{tabular}
\caption{\label{tab:stats_mono}Dataset statistics
         (source and number of tokens) for each language
         including both Indo-European and
         non-Indo-European languages.
 }
\end{table}

\paragraph{Monolingual Word Embeddings}
All monolingual embeddings  are
trained using a skip-gram model with negative
sampling~\citep{NIPS2013_5021}.  The dimension size is $100$ or
$200$.
All other hyperparameters are default in
Gensim~\cite{rehurek_lrec}.
 News articles except for Amharic are from Leipzig Corpora~\citep{Goldhahn12buildinglarge}.
 For Amharic, we use
documents from \textsc{lorelei}~\citep{LORELEI_lang_packs}.
MeCab \citep{kudo-yamamoto-matsumoto:2004:EMNLP} tokenizes
Japanese sentences.

\paragraph{Bilingual Seed Lexicon} For supervised methods, bilingual
lexicons from \citet{rolston2016collection} induce all cross-lingual
embeddings except for Danish, which uses Wiktionary.\footnote{\url{https://en.wiktionary.org/}}

\subsection{Cross-Lingual Mapping Algorithms}
\label{sec:algorithms}

We use three supervised (\abr{mse}, \abr{mse}+Orth, \abr{cca}) and
two unsupervised (\abr{muse}, \abr{vecmap}) cross-lingual
mappings:\footnote{We use the implementations
  from original authors with default parameters unless otherwise
  noted.}

\paragraph{Mean-squared error (\abr{mse})}
\citet{DBLP:journals/corr/MikolovLS13} minimize the mean-squared error
of bilingual entries in a seed lexicon to learn a projection between
two embeddings. We use the implementation by
\citet{artetxe2016learning}.

\paragraph{\abr{mse} with orthogonal constraints (\abr{mse}+Orth)}
\citet{xing-EtAl:2015:NAACL-HLT} add length normalization
and orthogonal constraints to preserve the
cosine similarities in the original monolingual embeddings.
\citet{artetxe2016learning} further preprocess monolingual embeddings
by mean centering.\footnote{One round of
iterative  normalization~\cite{iternorm}}

\paragraph{Canonical Correlation Analysis (\abr{cca})}
\citet{faruqui-dyer:2014:EACL}
maps two
monolingual embeddings into a shared space by maximizing the
correlation between translation pairs in a seed lexicon.

\paragraph{\citet[\abr{muse}]{lample2018word}}
use language-adversarial learning~\citep{ganin} to induce the initial bilingual seed lexicon, followed by a refinement step, which iteratively
solves the orthogonal Procrustes problem~\cite{Schonemann1966,artetxe-labaka-agirre:2017:Long}, aligning embeddings without an external
bilingual lexicon.
Like \abr{mse}+Orth, vectors are unit length and mean
centered.
Since \textsc{muse} is unstable~\cite{self_learn,eigenval_sim}, we
report the best of five runs.

\paragraph{\citet[\abr{vecmap}]{self_learn}}
induce an initial bilingual
seed lexicon by aligning intra-lingual similarity matrices computed
from each monolingual embedding.
We report the best of five runs to address
uncertainty from the initial dictionary.

\subsection{Modularity Implementation}
We implement modularity using random projection trees~\citep{dasgupta2008random} to speed up the
extraction of $k$-nearest
neighbors,\footnote{\url{https://github.com/spotify/annoy}} tuning
$k=3$ on the German \abrcamel{rcv}{2} dataset (Section~\ref{sec:k_sensitivity}).


\subsection{Task 1: Document Classification}
\label{sec:cldc}

We now explore the correlation of modularity and accuracy on cross-lingual document classification.
We classify documents from the Reuters \abrcamel{rcv}{1} and \abrcamel{rcv}{2} corpora~\citep{Lewis:2004:RNB:1005332.1005345}.
Documents have one of four labels (\underline{Corporate/Industrial},
\underline{Economics}, \underline{Government/Social}, \underline{Markets}).  We follow
\citet{klementiev-titov-bhattarai:2012:PAPERS}, except we use
all \textsc{en} training documents
and documents in each target language (\abr{da}, \abr{es}, \abr{it}, and \abr{ja}) as tuning and test data.
After removing out-of-vocabulary words,
we split documents in target languages into $10\%$ tuning data and $90\%$ test data.
Test data are 10,067 documents for \textsc{da}, 25,566
for \textsc{it}, 58,950 for \textsc{ja}, and 16,790 for \textsc{es}.
We exclude languages Reuters lacks: \textsc{hu} and \textsc{am}.
We use deep averaging
networks~\cite[\abr{dan}]{iyyer-EtAl:2015:ACL-IJCNLP} with three
layers, 100 hidden states, and 15 epochs as our classifier.
The \abr{dan} had better accuracy than averaged
  perceptron~\citep{avg_percep} in \citet{klementiev-titov-bhattarai:2012:PAPERS}.

\begin{figure}[tb]
\centering
     {\includegraphics[width=.95\linewidth]{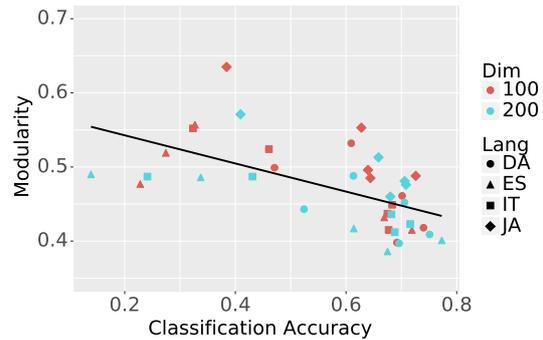}}
    \vspace{-2ex}
     \caption{\label{fig:mod_results} Classification accuracy
     and modularity of cross-lingual word embeddings ($\rho=-0.665$): less
     modular cross-lingual mappings have higher accuracy.
}
\end{figure}

\begin{table}[!tb]
  \small
  \centering
  \begin{tabular}{clrr}
                              & \bf Method           & \bf Acc. & \bf Modularity\\ \hline
                              &     \abr{mse}              & 0.399    & 0.529 \\
Supervised                    &     \abr{cca}              & 0.502    & 0.513 \\
                              &     \abr{mse}+Orth         & 0.628    & 0.452 \\ \hline
\multirow{2}{*}{Unsupervised} &     \textsc{muse}    & 0.711    & 0.431 \\
                              &     \textsc{vecmap}  & 0.643    & 0.432 \\
\hline
  \end{tabular}
 \caption{\label{tab:mod_results} Average classification accuracy on
   (\textsc{en} $\rightarrow$ \textsc{da}, \textsc{es}, \textsc{it},
   \textsc{ja}) along with the average modularity of five 
   cross-lingual word embeddings.
   \textsc{muse}
   has the best accuracy, captured by its low modularity.
 }
\end{table}

\paragraph{Results}

We report the correlation value computed from the data points in Figure~\ref{fig:mod_results}.
Spearman's correlation between modularity and classification
accuracy on all languages is $\rho = -0.665$.
Within each language pair,
modularity has a strong correlation within
\textsc{en}-\textsc{es} embeddings ($\rho=-0.806$),
\textsc{en}-\textsc{ja} ($\rho=-0.794$),
\textsc{en}-\textsc{it} ($\rho=-0.784$), and
a moderate correlation within
\textsc{en}-\textsc{da} embeddings ($\rho=-0.515$).
\textsc{muse} has the best classification accuracy (Table~\ref{tab:mod_results}),
reflected by its low modularity.

\begin{table}[tb]
\vspace{-1ex}
 \small
 \centering
  \begin{tabular}{|l|l|l|l|p{2cm}|p{2.5cm}|p{2.5cm}|}
  \hline
 \begin{CJK}{UTF8}{ipxm}市場 ``market''      \end{CJK}&\begin{CJK}{UTF8}{ipxm}終値\end{CJK} ``closing price'' \\\hline\hline
 \begin{CJK}{UTF8}{ipxm}新興 ``new coming''  \end{CJK}&\begin{CJK}{UTF8}{ipxm}上げ幅 ``gains''         \end{CJK}\\
 \begin{CJK}{UTF8}{ipxm}market               \end{CJK}&\begin{CJK}{UTF8}{ipxm}株価 ``stock price''     \end{CJK}\\
 \begin{CJK}{UTF8}{ipxm}markets              \end{CJK}&\begin{CJK}{UTF8}{ipxm}年初来 ``yearly''        \end{CJK}\\
 \begin{CJK}{UTF8}{ipxm}軟調 ``bearish''     \end{CJK}&\begin{CJK}{UTF8}{ipxm}続落 ``continued fall''  \end{CJK}\\
 \begin{CJK}{UTF8}{ipxm}マーケット ``market''\end{CJK}&\begin{CJK}{UTF8}{ipxm}月限 ``contract month''  \end{CJK}\\
 \begin{CJK}{UTF8}{ipxm}活況 ``activity''    \end{CJK}&\begin{CJK}{UTF8}{ipxm}安値 ``low price''       \end{CJK}\\
 \begin{CJK}{UTF8}{ipxm}相場 ``market price''\end{CJK}&\begin{CJK}{UTF8}{ipxm}続伸 ``continuous rise'' \end{CJK}\\
 \begin{CJK}{UTF8}{ipxm}底入 ``bottoming''   \end{CJK}&\begin{CJK}{UTF8}{ipxm}前日 ``previous day''    \end{CJK}\\
 \begin{CJK}{UTF8}{ipxm}為替 ``exchange''    \end{CJK}&\begin{CJK}{UTF8}{ipxm}先物 ``futures''         \end{CJK}\\
 \begin{CJK}{UTF8}{ipxm}ctoc                 \end{CJK}&\begin{CJK}{UTF8}{ipxm}小幅 ``narrow range''    \end{CJK}\\
  \hline
  \end{tabular}
 \caption{\label{tab:jp_nn_example} Nearest neighbors
 in an \textsc{en}-\textsc{ja} embedding.
 Unlike the \textsc{ja} word ``market'', the \textsc{ja} word ``closing price'' has no \textsc{en} vector nearby.
   }
\end{table}

\paragraph{Error Analysis}
A common error in \textsc{en} $\rightarrow$ \textsc{ja}
classification is predicting
\underline{Corporate/Industrial} for documents labeled \underline{Markets}.
One cause is documents with \begin{CJK}{UTF8}{ipxm}終値\end{CJK} ``closing price'';
 this has few market-based English neighbors
  (Table~\ref{tab:jp_nn_example}). \jbgcomment{Needs a little more
    explanation.  What's wrong with the nearest neighbors?}
  As a result, the model fails to transfer across languages.

\subsection{Task 2: Bilingual Lexical Induction (\abr{bli})}
\label{sec:task2}
Our second downstream task explores the correlation between modularity and
bilingual lexical induction (\abr{bli}).
We evaluate on the test set from \citet{lample2018word}, but we remove
pairs in the seed lexicon from \citet{rolston2016collection}.
The result is 2,099 translation pairs for \textsc{es}, 1,358 for
\textsc{it}, 450 for \textsc{da}, and 973 for \textsc{ja}.
We report precision@1 (P@1) for retrieving cross-lingual nearest
neighbors by cross-domain similarity local
scaling~\cite[\abr{csls}]{lample2018word}.

\begin{figure}[tb]
\centering
     {\includegraphics[width=.95\linewidth]{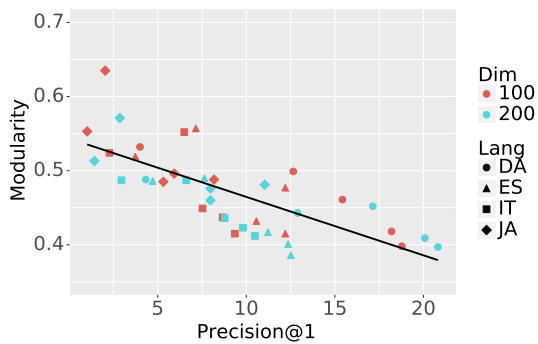}}
    \vspace{-2ex}
     \caption{\label{fig:bli_results} Bilingual lexical induction
       results and modularity of cross-lingual word embeddings
       ($\rho=-0.789$): lower modularity means higher precision@1.  }
\end{figure}

\paragraph{Results}
Although this task ignores intra-lingual nearest neighbors when
retrieving translations,
modularity still has a high correlation ($\rho=-0.785$) with P@1
(Figure~\ref{fig:bli_results}).
\textsc{muse} and \textsc{vecmap} beat the three supervised methods,
which have the lowest modularity (Table~\ref{tab:bli_mod_results}).
P@1 is low compared to other work on the \abr{muse} test set
(e.g., \citet{lample2018word}) because we filter out translation pairs which appeared
in the large training lexicon compiled by \citet{rolston2016collection},
and the raw corpora used to train monolingual embeddings (Table~\ref{tab:stats_mono})
are relatively small compared to Wikipedia.

\begin{table}[!tb]
  \small
  \centering
  \begin{tabular}{clrrr}
                               &  \bf Method     & \bf P@1 & \bf Modularity\\ \hline
                               &  \abr{mse}            & 7.30           &  0.529        \\
 Supervised                    &  \abr{cca}            & 3.06           &  0.513        \\
                               &  \abr{mse}+Orth       & 10.57          &  0.452        \\\hline
 \multirow{2}{*}{Unsupervised} &  \textsc{muse}  & 11.83          &  0.431        \\
                               &  \textsc{vecmap}& 12.92          &  0.432        \\
\hline
  \end{tabular}
  \vspace{-1ex}
 \caption{\label{tab:bli_mod_results} Average precision@1 on
   (\textsc{en} $\rightarrow$ \textsc{da}, \textsc{es}, \textsc{it},
   \textsc{ja}) along with the average modularity of the
   cross-lingual word embeddings trained with different methods.
   \textsc{vecmap} scores the best P@1, which is captured by its low modularity.
 }
\end{table}

\subsection{Task 3: Document Retrieval in Low-Resource Languages}
\label{sec:task3}

As a third downstream task, we turn to an important task for low-resource languages: lexicon
expansion~\citep{gupta-manning:2015:NAACL-HLT,hamilton-EtAl:2016:EMNLP2016} for document retrieval.
Specifically, we start with a set of \textsc{en}
seed words relevant to a particular concept, then 
find related words in a target language for which a
comprehensive bilingual lexicon does not exist.  We
focus on the disaster domain, where events may require immediate
{\abr{nlp}} analysis
(e.g., sorting \abr{sms}
messages to first responders).

We induce keywords in a target language by taking the $n$ nearest
neighbors of the English seed words in a cross-lingual word embedding.
We manually select sixteen disaster-related English seed words
from
Wikipedia articles, ``{\it Natural hazard}'' and ``{\it Anthropogenic
  hazard}''.  Examples of seed terms include ``earthquake'' and ``flood''.
Using the extracted terms, we retrieve disaster-related documents by
keyword matching and assess the coverage and relevance of terms by
area under the precision-recall curve (\abr{auc}) with varying $n$.

\vspace{-0.5ex}
\paragraph{Test Corpora}

As positively labeled documents, we use documents from the
\textsc{lorelei} project~\citep{LORELEI_lang_packs} containing any
disaster-related annotation.
There are $64$ disaster-related documents in Amharic, and $117$ in Hungarian.
We construct a set of negatively labeled documents from the Bible;
because the \textsc{lorelei} corpus does not include negative
documents and the Bible is available in all our
languages~\citep{Christodouloupoulos:2015},
we take the chapters of the gospels ($89$ documents), which do not discuss disasters,
and treat these as non-disaster-related documents.

\begin{table}[!tb]
     \small
     \centering
      \vspace{-1ex}
     \begin{tabular}{llrrrr}
       \bf Lang.                  & \bf Method      & \bf AUC &\bf Mod.\\ \hline
    \multirow{5}{*}{ \textsc{am}} 
                                  & \abr{mse}             & 0.578  & 0.628 \\
                                  & \abr{cca}             & 0.345  & 0.501 \\
                                  & \abr{mse}+Orth        & 0.606  & 0.480 \\
                                  & \textsc{muse}   & 0.555  & 0.475 \\
                                  & \textsc{vecmap} & 0.592  & 0.506 \\ \hline
    \multirow{5}{*}{ \textsc{hu}} 
                                  & \abr{mse}             & 0.561  & 0.598 \\
                                  & \abr{cca}             & 0.675  & 0.506 \\
                                  & \abr{mse}+Orth        & 0.612  & 0.447 \\
                                  & \textsc{muse}   & 0.664  & 0.445 \\
                                  & \textsc{vecmap} & 0.612  & 0.432 \\ \hline
    \multicolumn{2}{c}{Spearman Correlation $\rho$} & \multicolumn{2}{c}{$-0.378$}  \\
     \end{tabular}
     \vspace{-1ex}
    \caption{\label{tab:prec_results}
    Correlation between modularity and AUC on document retrieval.
    It shows a moderate correlation to this task.
      }
\end{table}

\paragraph{Results}
Modularity has a moderate correlation with \abr{auc}
($\rho=-0.378$, Table~\ref{tab:prec_results}).
While modularity focuses on the entire vocabulary of cross-lingual
word embeddings, this task focuses on a small, specific
subset---disaster-relevant words---which may explain the low
correlation compared to \abr{bli} or document
classification.

\section{Use Case: Model Selection for \textsc{muse}}
\label{sec:validation}

A common use case of intrinsic measures is model selection.
We focus on \textsc{muse}~\cite{lample2018word} since it is unstable,
especially on distant language
pairs~\cite{self_learn,eigenval_sim,non_adv} and therefore requires an
effective metric for model selection.
\textsc{muse} uses a validation metric in its two steps: (1) the language-adversarial step, and (2) the refinement step.
First the algorithm selects an optimal mapping $W$ using a validation metric, obtained from language-adversarial learning~\cite{ganin}.
Then the selected mapping $W$ from the language-adversarial step is passed on to the refinement step~\citep{artetxe-labaka-agirre:2017:Long} to re-select the optimal mapping $W$ using the same validation metric after each epoch of solving the orthogonal Procrustes problem~\citep{Schonemann1966}.

Normally, \textsc{muse} uses an intrinsic metric, \textsc{csls} of the top 10K frequent words \cite[\textsc{csls}-10K]{lample2018word}.  Given word vectors $s, t \in \mathbb{R}^n$ from a source and a target embedding, \textsc{csls} is a cross-lingual similarity metric,
\begin{equation}
\text{\textsc{csls}}(W s, t) = 2 \cos(W s, t) - r(W s) - r(t)
\end{equation}
where $W$ is the trained mapping after each epoch, and $r(x)$ is the average cosine similarity of the top $10$ cross-lingual nearest neighbors of a word $x$.

What if we use modularity instead?
To test modularity as a validation metric for~\abr{muse}, we
compute modularity on the lexical graph of 10K most frequent words
(Mod-10K; we use 10K for consistency with \textsc{csls} on the same
words) after each epoch of the adversarial step and the refinement step and select the best mapping.

The important difference between these two metrics is that Mod-10K
considers the relative similarities between intra- and cross-lingual
neighbors, while \textsc{csls}-10K only considers the similarities of
cross-lingual nearest neighbors.\footnote{Another difference is that
  $k$-nearest neighbors for \textsc{csls}-10K is $k=10$, whereas
  Mod-10K uses $k=3$. However, using $k=3$ for \textsc{csls}-10K leads
  to worse results; we therefore only report the result on the
  default metric i.e., $k=10$.}

\begin{table}[!tb]
     \small
     \centering
     \begin{tabular}{llrrrrrrrr}
 \bf Family                      & \bf Lang.   & \multicolumn{2}{c}{\bf \textsc{csls}-10K} &  \multicolumn{2}{c}{\bf Mod-10K}   \\ \hline
                                 &             & Avg.           & Best                     & Avg.         & Best    \\ \hline
\multirow{2}{*}{Germanic}        & \textsc{da} &  \bf 52.62      &  \bf 60.27               & 52.18        & 60.13   \\
                                 & \textsc{de} &  \bf 75.27      &  \bf 75.60               & 75.16        & 75.53   \\ \hline
\multirow{2}{*}{Romance}         & \textsc{es} & \bf 74.35       &  83.00                   & 74.32        & 83.00   \\
                                 & \textsc{it} &   78.41         &  78.80                   & \bf 78.43    & 78.80   \\ \hline
\multirow{3}{0.7cm}{Indo-Iranian}& \textsc{fa} &  \bf 27.79      & 33.40                    & 27.77        & 33.40   \\
                                 & \textsc{hi} &  25.71          & 33.73                    & \bf 26.39    & \bf 34.20\\
                                 & \textsc{bn} &  0.00           & 0.00                     & \bf 0.09     & \bf 0.87 \\ \hline
\multirow{8}{*}{Others}          & \textsc{fi} &  4.71           & 47.07                    & 4.71         & 47.07    \\
                                 & \textsc{hu} & \bf 52.55       & 54.27                    & 52.35        & \bf 54.73\\
                                 & \textsc{ja} &  18.13          & 49.69                    & \bf 36.13    & 49.69     \\
                                 & \textsc{zh} &  5.01           & 37.20                    & \bf 10.75    & 37.20     \\
                                 & \textsc{ko} &  16.98          & 20.68                    & \bf 17.34    & \bf 22.53 \\
                                 & \textsc{ar} &  15.43          & 33.33                    & \bf 15.71    & \bf 33.67 \\
                                 & \textsc{id} &  67.69          & 68.40                    & \bf 67.82    & 68.40     \\
                                 & \textsc{vi} &  0.01           & 0.07                     & 0.01         & 0.07      \\ \hline
     \end{tabular}
    \caption{\label{tab:validation_metric} \abr{bli} results (P@1
      $\times 100\%$) from \textsc{en} to each target language with
      different validation metrics for \textsc{muse}: default
      (\textsc{csls}-10K) and modularity (Mod-10K).  We report the
      average (Avg.) and the best (Best) from ten runs with ten random
      seeds for each validation metric.  \textbf{Bold} values are
      mappings that are not shared between the two validation metrics.
      Mod-10K improves the robustness of \textsc{muse} on distant
      language pairs.
      }
\end{table}

\paragraph{Experiment Setup}

We use the pre-trained
fastText vectors~\citep{fasttext} to be comparable with the prior
work.  Following \citet{self_learn}, all vectors are unit length
normalized, mean centered, and then unit length
normalized.
We use the test lexicon by \citet{lample2018word}.
We run ten times with the same random seeds and
hyperparameters but with different validation metrics.  Since
\textsc{muse} is unstable on distant language
pairs~\cite{self_learn,eigenval_sim,non_adv}, we test it on
English to languages from diverse language families: Indo-European
languages such as Danish (\textsc{da}), German (\textsc{de}), Spanish
(\textsc{es}), Farsi (\textsc{fa}), Italian (\textsc{it}), Hindi
(\textsc{hi}), Bengali (\textsc{bn}), and non-Indo-European languages
such as Finnish (\textsc{fi}), Hungarian (\textsc{hu}), Japanese
(\textsc{ja}), Chinese (\textsc{zh}), Korean (\textsc{ko}), Arabic
(\textsc{ar}), Indonesian (\textsc{id}), and Vietnamese (\textsc{vi}).

 \paragraph{Results}
Table~\ref{tab:validation_metric} shows P@1 on \abr{bli} for each target
language using English as the source language.
Mod-10K improves P@1 over the default validation metric in diverse
languages, especially on the average P@1 for non-Germanic languages
such as \textsc{ja} ($+18.00\%$) and \textsc{zh} ($+5.74\%$), and the
best P@1 for \textsc{ko} ($+1.85\%$).
These language pairs include pairs (\textsc{en-ja} and
\textsc{en-hi}), which are difficult for \textsc{muse}~\cite{non_adv}.
Improvements in \textsc{ja} come from selecting a better mapping
 during the refinement step, which the default validation
misses.  For \textsc{zh}, \textsc{hi}, and
\textsc{ko}, the improvement comes from selecting better mappings
during the adversarial step.
However, modularity does not improve
on all languages (e.g., \textsc{vi}) that are reported to fail by \citet{non_adv}.

\section{Analysis: Understanding Modularity as an Evaluation Metric}
\label{sec:empirical}

The experiments so far 
show that modularity captures whether an
embedding is useful,
which suggests that modularity could be used as an intrinsic evaluation or validation metric.
Here, we investigate whether modularity can
capture \emph{distinct} information compared to existing evaluation measures:
\textsc{qvec-cca}~\cite{ammar2016massively}, \textsc{csls}~\cite{lample2018word}, and cosine similarity between translation pairs (Section~\ref{sec:ablation}).
We also 
analyze the effect of
the number of nearest neighbors $k$ (Section~\ref{sec:k_sensitivity}).

\begin{figure}[tb]
 \centering
     {\includegraphics[width=0.98\linewidth]{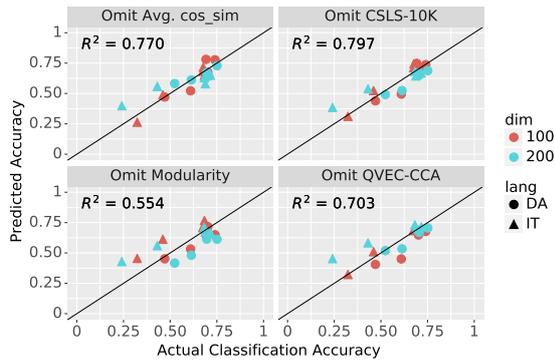}}
    \vspace{-1ex}
     \caption{We predict the cross-lingual document classification results for \textsc{da} and \textsc{it} from Figure~\ref{fig:mod_results} using three out of four evaluation metrics.  Ablating modularity causes by far the largest decrease ($R^2 = 0.814$ when using all four features) in $R^2$, showing that it captures
       information complementary to the other metrics.
       }
\label{fig:ablation}
\end{figure}

\subsection{Ablation Study Using Linear Regression}
\label{sec:ablation}

We fit a linear regression model to predict the classification
accuracy given four intrinsic measures: \textsc{qvec-cca}, \textsc{csls}, average cosine
similarity of translations, and modularity.  We ablate 
each of the four measures, fitting linear regression with standardized feature values,
for two target languages (\textsc{it} and \textsc{da}) on the task of cross-lingual document classification (Figure~\ref{fig:mod_results}).  We limit to
\textsc{it} and \textsc{da} because aligned supersense annotations to
\textsc{en} ones~\citep{Miller:1993:SC:1075671.1075742}, required for \textsc{qvec-cca} are only available
in those languages~\citep{Montemagni2003,martinezalonso-EtAl:2015:NODALIDA,martinezalonsoetal2016,ammar2016massively}.
We standardize the values of the four features before training the regression model.

Omitting modularity hurts accuracy prediction on
cross-lingual document classification substantially, while omitting
the other three measures has smaller effects
(Figure~\ref{fig:ablation}).  Thus, modularity complements the
other measures and is more predictive of classification
accuracy.

\subsection{Hyperparameter Sensitivity}
\label{sec:k_sensitivity}

While modularity itself does not have any adjustable hyperparameters,
our approach to constructing the lexical graph has two hyperparameters: the number of nearest neighbors ($k$) and the number of trees ($t$) for approximating the $k$-nearest neighbors using random projection trees.
We conduct a grid search for $k \in \{1, 3, 5, 10, 50, 100, 150, 200\}$ and $t \in \{50, 100, 150, 200, 250, 300, 350, 400, 450, 500\}$ using the German \abrcamel{rcv}{2} corpus as the held-out language to tune hyperparameters.

\begin{figure}[t]
   \centering
      \vspace{-2ex}
       {\includegraphics[width=0.85\linewidth]{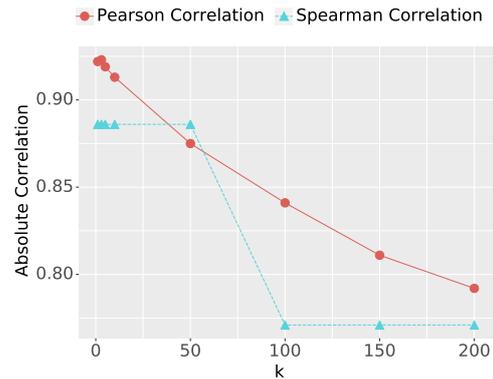}}
      \vspace{-1ex}
       \caption{\label{fig:diff_k_corr} Correlation between modularity and classification performance (\textsc{en}$\rightarrow$\textsc{de}) with different numbers of neighbors $k$. 
       Correlations are computed on the same setting as Figure~\ref{fig:mod_results} using supervised methods. 
       We use this to set $k=3$. 
       }
\end{figure}

The nearest neighbor $k$ has a much larger effect on modularity than~$t$, 
so we focus on analyzing the effect of $k$, 
using the optimal $t=450$.
Our earlier experiments all use $k=3$ since it gives the highest Pearson's and Spearman's correlation on the tuning dataset (Figure~\ref{fig:diff_k_corr}). 
The absolute correlation between the downstream task decreases when setting $k>3$, indicating nearest neighbors beyond $k=3$ are only contributing noise.

\section{Discussion: What Modularity Can and Cannot Do}
\label{sec:conc}

This work focuses on modularity as a diagnostic tool: it is cheap and
effective at discovering which embeddings are likely to falter on
downstream tasks.  
Thus, practitioners should consider including it as a metric for
evaluating the quality of their embeddings.
Additionally, we believe that modularity could serve as a useful prior for the
algorithms that \emph{learn} cross-lingual word embeddings: during learning
prefer updates that avoid increasing modularity if all else is equal.

Nevertheless, we recognize limitations of modularity. 
Consider the following cross-lingual word embedding ``algorithm'': for
each word, select a random point on the unit hypersphere.
This is a horrible distributed representation: the position
of words' embedding has no relationship to the underlying meaning.
Nevertheless, this representation will have very low modularity.
Thus, while modularity can identify bad embeddings, once vectors are
well mixed, this metric---unlike \textsc{qvec} or \textsc{qvec-cca}---cannot identify whether the
meanings make sense.
Future work should investigate how to combine techniques that use both
word meaning and nearest neighbors for a more robust, semi-supervised
cross-lingual evaluation.

\section*{Acknowledgments}
This work was supported by \abr{nsf} grant \abr{iis}-1564275
and by \abr{darpa} award HR0011-15-C-0113 under subcontract to Raytheon \abr{bbn} Technologies.
The authors would like to thank Sebastian Ruder, Akiko Aizawa,
the members of the \abr{clip} lab at the University of Maryland,
the members of the \abr{clear} lab at the University of Colorado,
and the anonymous reviewers for their feedback.
The authors would like to also thank Mozhi Zhang for
providing the deep averaging network code.
Any opinions, findings, conclusions, or recommendations expressed here
are those of the authors and do not necessarily reflect the view of
the sponsor.

\bibliography{2019_acl_modularity}
\bibliographystyle{acl_natbib}

\end{document}